\documentclass[11pt]{article}
\usepackage{scrextend}
\usepackage{style}
\usepackage{times}
\usepackage{url}
\usepackage{latexsym}
\usepackage{graphicx}
\usepackage{caption}
\usepackage{enumitem}
\usepackage{subfig}
\usepackage{appendix}
\usepackage{amssymb}
\usepackage{amsmath}
\usepackage{svg}
\usepackage{multirow}
\usepackage{float}
\usepackage{hyperref}
\usepackage{tablefootnote}
\setsvg{inkscape = inkscape -z -D}
\setsvg{svgpath = figs/}
\eaclfinalcopy

\title{Measuring Conversational Fluidity in Automated Dialogue Agents}

\author{
  {\bf Keith Vella$^1$, Massimo Poesio$^1$, Michael Sigamani$^2$, Cihan Dogan$^2$, Aimore Dutra$^2$}, \\ 
  {\bf Dimitrios Dimakopoulos$^2$, Alfredo Gemma$^2$, Ella Walters$^2$}\\
  $^1$Queen Mary College, University of London \\
  $^2$Constellation AI \\
}

\date{}
\begin{document}
\maketitle
\begin{abstract}
We present an automated evaluation method to
measure fluidity in conversational dialogue systems.
The method combines various state of the art Natural Language tools
into a classifier, and human ratings on these dialogues to train an automated
judgment model. Our experiments show that
the results are an improvement on existing metrics for measuring fluidity.
\end{abstract}

\section{Introduction}

Conversational interactions between humans and Artificial Intelligence (AI)
agents could amount to as much as thousands of interactions
a day given recent developments~\cite{bench2007argumentation}. 
This surge in human-AI interactions has led to an interest in developing more fluid interactions between agent and human.
The term `fluidity', when we refer to dialogue systems, tries to measure the concept of how humanlike
communication is between a human and an AI entity.
Conversational fluidity has historically been measured using metrics such as perplexity, recall, and F1-scores.
However, one finds various drawbacks using these metrics.
During the automatic evaluation stage of the second Conversational Intelligence Challenge (ConvAI2)~\cite{dinansecond2019} competition,
it was noted that consistently replying with ``I am you to do and your is like''
would outperform the F1-score of all the models in the competition.
This nonsensical phrase was constructed simply by picking several frequent words from the training set.
Also, Precision at K, or the more specific Hits@1 metric has been used historically in assessing retrieval based aspects of the agent.
This is defined as the accuracy of the next dialogue utterance when choosing
between the gold response and N--1 distractor responses.
Since these metrics are somewhat flawed, human evaluations were used in conjunction.
Multiple attempts have been made historically to try to develop automatic metrics to assess dialogue fluidity.
One of the earliest Eckert et al. (1997), used a stochastic system which regulated user-generated
dialogues to debug and evaluate chatbots~\cite{658991}.
In the same year, Marilyn et al. (1997) proposed the
PARADISE~\cite{DBLP:journals/corr/cmp-lg-9704004} model.
This framework was developed to evaluate dialogue agents in spoken conversations.
A few years later the BLEU~\cite{papinenibleu2002} metric was proposed. Subsequently, for almost two decades,
this metric has been one of the few to be widely adopted
by the research community.
The method, which compares the matches in n-grams from the
translated outputted text and the input text proved to be quick,
inexpensive and has therefore been widely used.
Therefore, we use the BLEU metric as a baseline to compare the quality of our proposed model.

\subsection{Datasets}
\label{datasets}

For this study, we use two types of data namely single-turn and multi-turn.
The first type, single-turn, is defined such that each instance is made up of one statement and one response.
This pair is usually a fragment of a larger dialogue.
When given to humans for evaluation of fluidity, we ask to give a score on characteristics such as
``How related is the response to the statement?'' or ``Does the response contain repeated text from the user's statement?''.
These are all things that should not be affected by the fact that no history or context is provided and therefore,
can still be classified reasonably. Contrary to the single turn datasets, the second type is the multi-turn dataset.
This contains multiple instances of statements and responses, building on each other to create a fuller conversation.
With these kinds of datasets, one can also evaluate and classify the data on various other attributes.
An example of such evaluations would be something like ``Does this response continue on the flow of the conversation?''
or ``Is the chatbot using repetitive text from previous responses?''. The details of how we collected each dataset are detailed below.

\textbf{Single-Turn:}
This dataset consists of single-turn instances of statements and responses from the MiM chatbot developed at Constellation AI~\cite{constellationai}.
The responses provided were then evaluated using Amazon Mechanical Turk (AMT) workers.
A total of five AMT workers evaluated each of these pairs. The mean of the five evaluations is then used as the target variable.
A sample can be seen in Table~\ref{tab:replika_full}. This dataset was used during experiments with results published in the
Results section.~\footnote{The data is available and can be accessed here: \href{https://github.com/ConstellationAI/DialogueFluidity/blob/master/single\_turn.csv}{here}.}

\textbf{Multi-Turn:}
This dataset is taken from the ConvAI2 challenge and consists of various types of dialogue that have been generated by human-computer conversations.
At the end of each dialogue, an evaluation score has been given, for each dialogue, between 1--4.

\begin{table*}[htbp]
  \begin{center}
  \caption{A sample from the single-turn dataset. Each statement and response pair 
           has been rated by five different AMT workers, with the mean of the scores also shown.}
        \resizebox{\textwidth}{!}{
   \begin{tabular}{c|c|c|c|c|c|c|c}
      \textbf{Statement} & \textbf{Response} 
      & \textbf{AMT1} & \textbf{AMT2} & \textbf{AMT3}
      & \textbf{AMT4}& \textbf{AMT5}& \textbf{Mean}\\
      \hline
      ahahah i have got easily the most loyal pig ever & That's nice, hah. & 4 & 3 & 3 & 2 & 5 & 3.4\\
      i can bake you a cake for your birthday & Oh, I would really appreciate that. & 2 & 4 & 4 & 4 & 5 & 3.8\\
      i do too but my ginger snaps & Do you ever exaggerate your stories? & 1 & 1 & 3 & 3 & 1 & 1.8\\
      yes it can do you have other hobbies ? & All kinds, my taste is very eclectic. & 3 & 2 & 3 & 4 & 4 & 3.2\\
    \end{tabular}}
    \label{tab:replika_full}
  \end{center}
\end{table*}

\section{Method}

This section discusses the methods and used to develop our attributes and the 
technical details of how they are combined to create a final classification layer.

\subsection{BERT Next Sentence Prediction}

BERT~\cite{DBLPjournalscorrabs181004805} is a state-of-the-art model, which has been pre-trained on a
large corpus and is suitable to be fine-tuned for various downstream NLP tasks.
The main innovation between this model and existing language models is in how the model is trained.
For BERT, the text conditioning happens on both the left and right context of every word and is therefore bidirectional. In previous models~\cite{radford2018improving},
a unidirectional language model was usually used in the pre-training.
With BERT, two fully unsupervised tasks are performed.
The Masked Language Model and the Next Sentence Prediction (NSP).

\begin{figure*}[!htbp]
        \centering
        \includegraphics[width=0.49\textwidth]{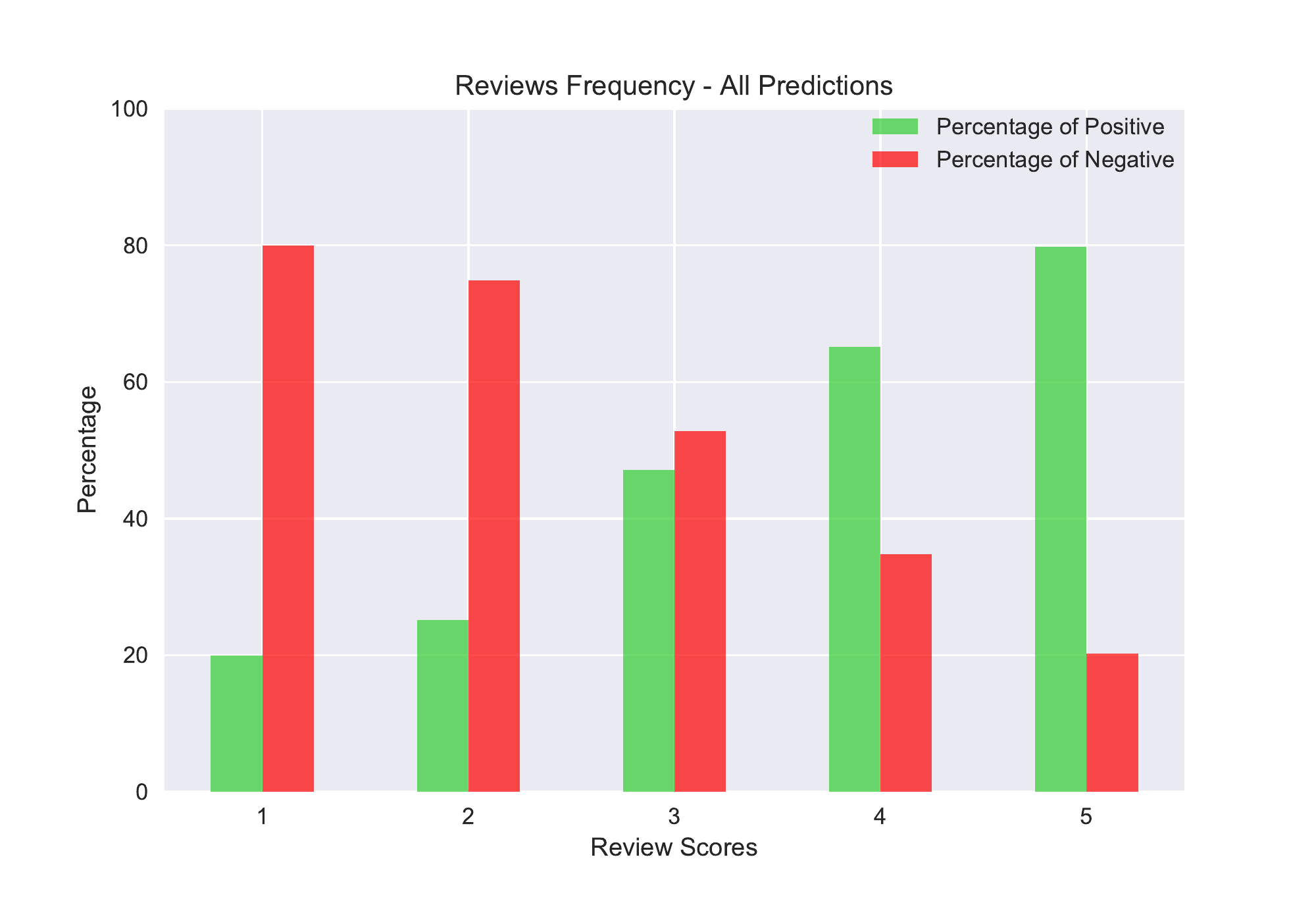}
        \includegraphics[width=0.49\textwidth]{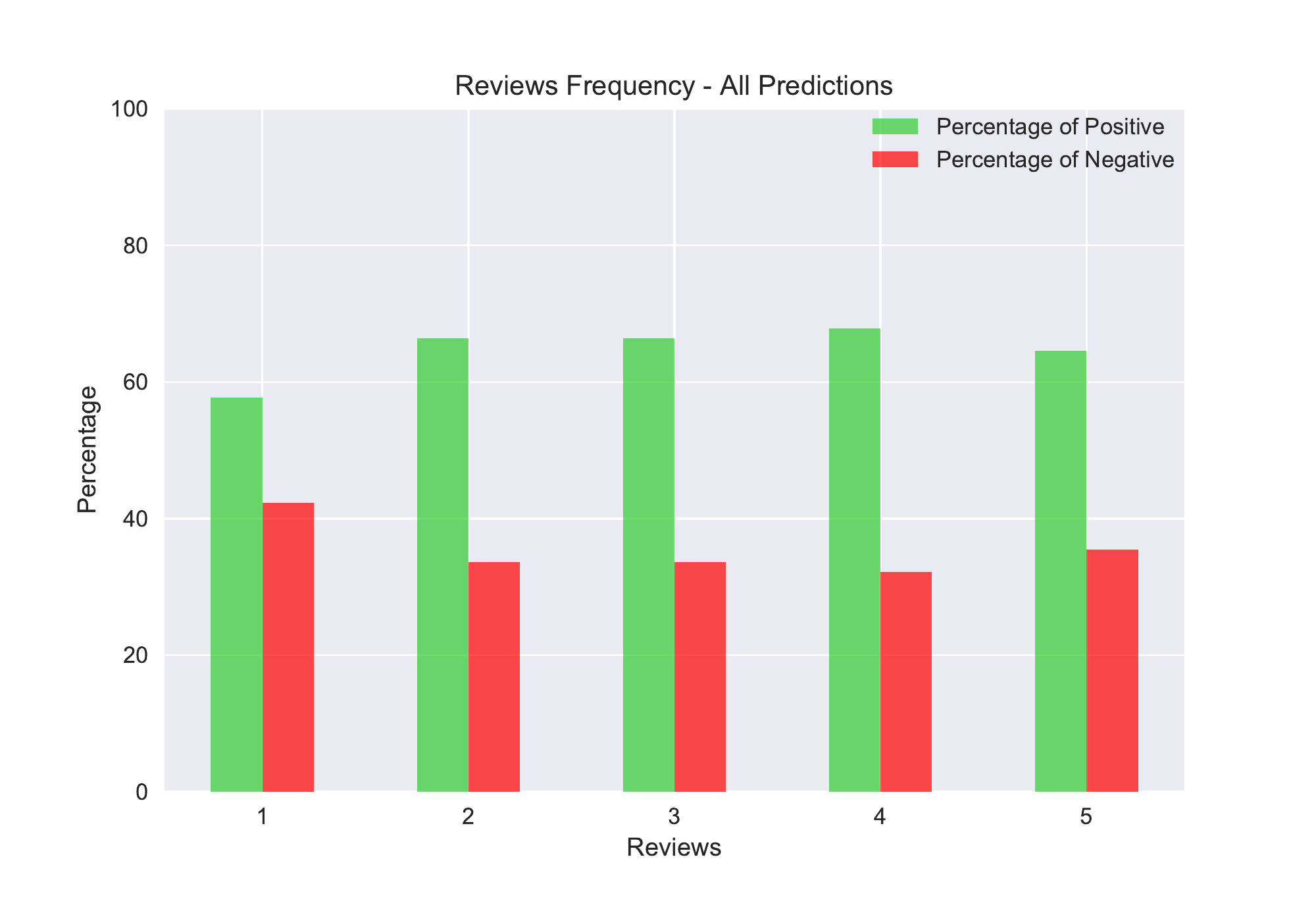}
        \caption{A histogram showing the BERT predictions for NSP on the single-turn (left) 
                 and multi-turn (right) datasets. We show the percentage of positive and negative predictions
                 for each category.}
        \label{fig:bertPredictions}
\end{figure*}

For this study, the NSP is used as a proxy for the relevance of response. 
Furthermore, in order to improve performance, we fine-tune on a customized dataset which achieved an accuracy of 82.4\%.
For the main analysis, we used the single-turn dataset,
which gave us a correlation of 0.28 between the mean of the AMT evaluation and the BERT NSP.
Next, we put each score into a category. For example, if the average score is 2.3, this would be placed in category 2.
We then displayed the percentage of positive and negative predictions in a histogram for each of the categories.
As seen in Figure~\ref{fig:bertPredictions}, a clear pattern is seen between the higher scores and the positive prediction,
and the lower scores and the negative predictions. details of how they are combined to create a final classification layer.

\subsection{Repetition Control}

This attribute is calculated by checking each statement and response for various types of repetition by using n-gram overlap. 
The motivation for including this as an attribute in dialogue fluidity is that repetitive words or n-grams 
can be bothersome to the end-user. 

Repetitions are measured according to whether they are internal, external or partner. 
We calculate a percentage based on the single-turn utterance or the entire multi-turn conversation. 
We use unigram, bigram, and trigram for each repetition type based off~\cite{seewhat2019}. 

We calculate a correlation of each repetition module with respect to human evaluations in order to understand the impact. 
For the single-turn dataset, the correlation is -0.09 and 0.07 for the internal and partner repetition attribute respectively.  
For the multi-turn dataset the correlation was -0.05 and -0.02 for the internal and partner repetition attribute respectively. 
This low correlation is reasonable and was expected. Measuring repetition in this way is not expected to provide huge classification power. 
However, we will attempt to exploit differences in correlation between these attributes and ones described below, which will provide some classification power. 

\subsection{Balance of Dialogue}

For this attribute, we calculated the number of questions asked. 
For this particular case, we are not able to measure a correlation with respect to human evaluations. 

\subsection{Short-Safe Answers}

\begin{figure*}[htpb]
\centering
\includegraphics[width=0.9\textwidth]{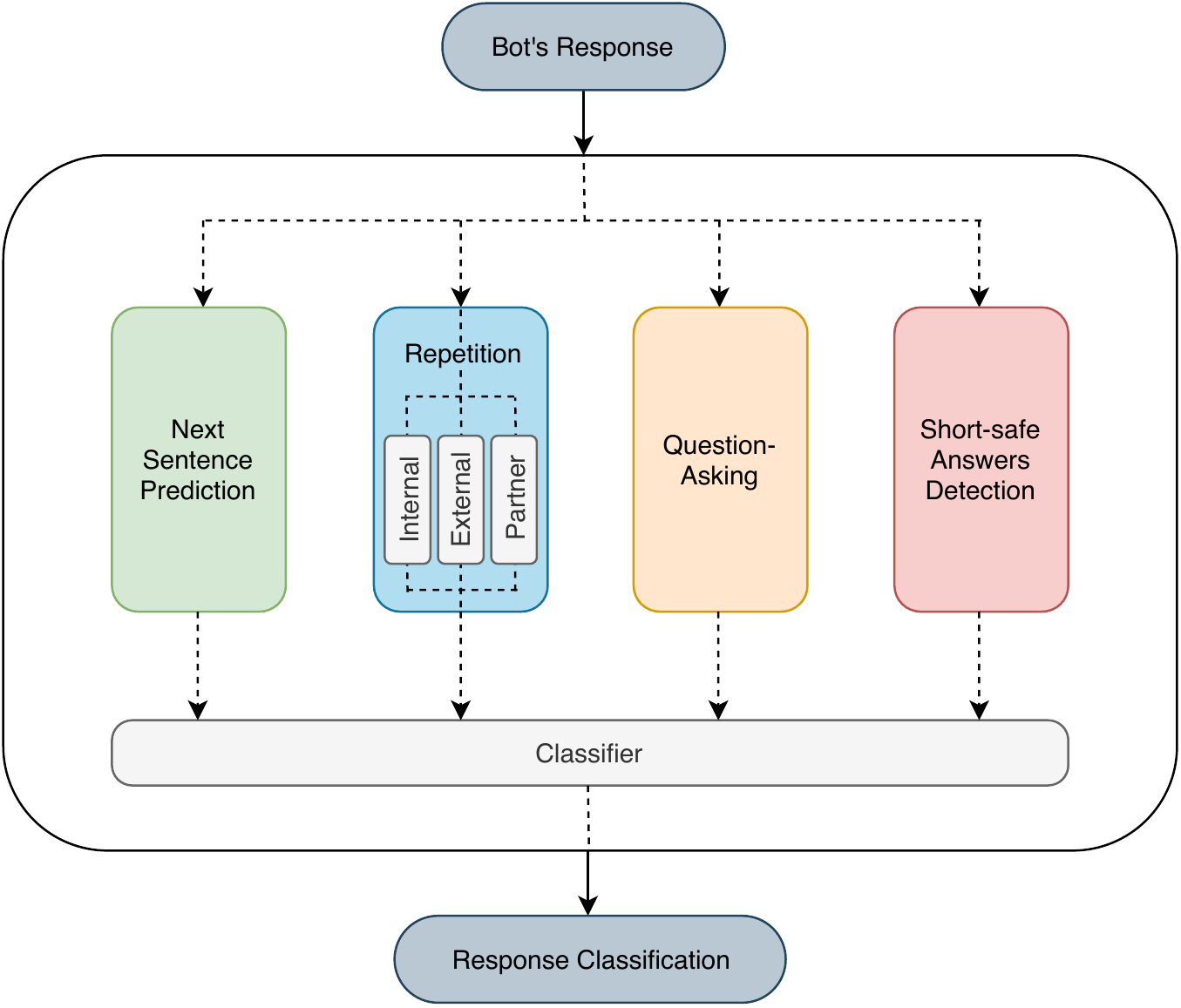}
\caption{An overview of the architecture of the combined module. 
         The agent's response is received as input to the component. 
         The data is passed through each attribute in a parallel fashion. 
         Each attribute will output either a float or an integer value, 
         which is then used to train a classifier.}
\label{fig:combinedModule} 
\end{figure*}

Here, we checked for the length of the utterance and the presence of a Named Entity.
We checked the correlation of this attribute with the human evaluation scores.
The correlation score attained on the single-turn dataset was -0.09,
while for the multi-turn dataset the correlation was 0.
The full pipeline can be seen diagrammatically in Figure~\ref{fig:combinedModule}.

\section{Results}

To create a final metric, we combine the individual components from Section~2
as features into a Support Vector Machine.       
The final results for our F1-score from this classification technique are 0.52 and 0.31 
for the single and multi-turn data respectively.

We compare our results for both the single-turn and multi-turn experiments 
to the accuracies on the test data based off the BLEU score. 
We see an increase of 6\% for our method with respect to the BLEU score in the single turn data, 
and a no change when using the multi-turn test set.

\section{Conclusion}

This study aimed to implement an automatic metric to assess the fluidity of dialogue systems.
We wanted to test if any set of attributes could show a high correlation to manual evaluations, 
thus replacing it entirely. As highlighted in the ConvAI2 challenge, 
automatic metrics are not reliable for a standalone evaluation of low-level dialogue outputs. 
For this study, three attributes were investigated.
Tests were carried out based on these proposed attributes by making use of single and multi-turn datasets. 
These attributes, combined with the BERT model, showed that our classifier performed 
better than the BLEU model for the single-turn dataset. 
However, no improvement was seen on the multi-turn dataset. 

Concerning feature importance, we observed that internal repetition 
and NSP are the most important attributes when used to classify fluidity. 
We believe that further work can be carried out in finding a more discriminating 
set of attributes.

\bibliography{eacl2017}

\begin{thebibliography}{21}
\expandafter\ifx\csname natexlab\endcsname\relax\def\natexlab#1{#1}\fi

\bibitem[{Bench-Capon TJ, Dunne PE.}]{bench2007argumentation}
Bench-Capon TJ, Dunne PE.
\newblock Argumentation in artificial intelligence.
\newblock Artificial intelligence. 2007;171(10-15):619--641.

\bibitem[\protect\citename{Dinan E, Logacheva V, Malykh V, 
Urbanek J, et~al.}]{dinansecond2019}
Dinan E, Logacheva V, Malykh V, Miller AH, Shuster K, Urbanek J, et~al.
\newblock The {Second} {Conversational} {Intelligence} {Challenge} ({ConvAI}2).
\newblock CoRR. 2019;abs/1902.00098.
\newblock Available from: \url{http://arxiv.org/abs/1902.00098}.

\bibitem[{{Eckert} W, {Levin} E, {Pieraccini} R.}]{658991}
Eckert W, Levin E, Pieraccini R.
\newblock User modeling for spoken dialogue system evaluation.
\newblock In: 1997 IEEE Workshop on Automatic Speech Recognition and
  Understanding Proceedings; 1997. p. 80--87.

\bibitem[{Walker MA, Litman DJ, Kamm CA, Abella A.}]{DBLP:journals/corr/cmp-lg-9704004}
\newblock {PARADISE:} {A} Framework for Evaluating Spoken Dialogue Agents.
\newblock CoRR. 1997;cmp-lg/9704004.
\newblock Available from: \url{http://arxiv.org/abs/cmp-lg/9704004}.


\bibitem[{Papineni K, Roukos S, Ward T, Zhu WJ.}]{papinenibleu2002}
\newblock BLEU: a method for automatic evaluation of machine translation.
\newblock In: Proceedings of the 40th annual meeting on association for
  computational linguistics. Association for Computational Linguistics; 2002.
  p. 311--318.
\bibitem[{Constellation AI}]{constellationai}
\newblock Available from: \url{https://constellation.ai}.


\bibitem[{Devlin J, Chang M, Lee K, Toutanova K.}]{DBLPjournalscorrabs181004805}
Devlin J, Chang M, Lee K, Toutanova K.
\newblock {BERT:} Pre-training of Deep Bidirectional Transformers for Language
  Understanding.
\newblock CoRR. 2018;abs/1810.04805.
\newblock Available from: \url{http://arxiv.org/abs/1810.04805}.


\bibitem[{Khatri C, Hedayatnia B, Nunn J, Pan Y, Liu Q, et~al.}]{DBLPjournalscorrabs181210757}
\newblock Advancing the State of the Art in Open Domain Dialog Systems through
  the Alexa Prize.
\newblock CoRR. 2018;abs/1812.10757.
\newblock Available from: \url{http://arxiv.org/abs/1812.10757}.


\bibitem[{Radford A, Narasimhan K, Salimans T, Sutskever I.}]{radford2018improving}
\newblock Improving language understanding by generative pre-training.
\newblock URL
  https://s3-us-west-2amazonawscom/openai-assets/research-covers/language-unsupervised/language\-understanding\-paperpdf.
  2018;.


\bibitem[{See A, Roller S, Kiela D, Weston J.}]{seewhat2019}
\newblock What makes a good conversation? {How} controllable attributes affect
  human judgments.
\newblock CoRR. 2019;abs/1902.08654.
\newblock Available from: \url{http://arxiv.org/abs/1902.08654}.


\bibitem[{DeepPavlov. The Conv. Intelligence Challenge 2.}]{convaidata}
\newblock Github Pages; 2019.
\newblock Date accessed: 17 August 2019.
\newblock Available from: \url{http://convai.io/data/}.

\end{thebibliography}
\bibliographystyle{eacl2017}

\end{document}